\documentclass[sigconf]{acmart}
\AtBeginDocument{%
  }

\copyrightyear{2026}
\acmYear{2026}
\setcopyright{cc}
\setcctype{by}

\acmConference[MM '26]
{Proceedings of the 34th ACM International Conference on Multimedia}
{November 10--14, 2026}
{Rio de Janeiro, Brazil}

\acmBooktitle{Proceedings of the 34th ACM International Conference on Multimedia
(MM '26), November 10--14, 2026, Rio de Janeiro, Brazil}

\acmDOI{10.1145/3767308.3834907}
\acmISBN{979-8-4007-2213-4/2026/11}

\usepackage{booktabs}
\usepackage{makecell}
\usepackage{multirow}
\usepackage{array}
\usepackage[table,xcdraw]{xcolor}
\usepackage{tikz}
\usepackage{amsfonts}
\usepackage{amsmath}
\usepackage{colortbl}
\usepackage{algorithm}
\usepackage{algorithmic}
\usepackage{enumitem}

\usepackage{pifont}

\newcommand{\coloredcircle}[2]{%
  \tikz[baseline=(char.base)]{
    \node[
      shape=circle,
      fill=#1,
      inner sep=1.0pt,
      minimum size=1.0em
    ] (char) {\textcolor{white}{#2}};
  }%
}

\DeclareRobustCommand{\captioncircle}[2]{%
  \tikz[baseline=(char.base)]{
    \node[
      shape=circle,
      fill=#1,
      inner sep=0.8pt,
      minimum size=1.0em
    ] (char) {\textcolor{white}{\footnotesize #2}};
  }%
}

\definecolor{myblue}{RGB}{52,152,219}
\definecolor{mygreen}{RGB}{46,204,113}
\definecolor{mypurple}{RGB}{155,89,182}

\newcommand{\cmark}{\textcolor{black}{\ding{51}}}
\newcommand{\xmark}{\textcolor{gray!60}{\ding{55}}}

\definecolor{lightblue}{HTML}{DCEFF8}
\definecolor{deepblue}{HTML}{C9E3F2}
\definecolor{lightpink}{HTML}{FDDAE2}

\begin{document}

%%
%% The "title" command has an optional parameter,
%% allowing the author to define a "short title" to be used in page headers.
\title{H²AL: Hyperbolic Hierarchy-aware Aggregative Learning for Registration-based Few-shot Medical Image Segmentation}

%%
%% The "author" command and its associated commands are used to define
%% the authors and their affiliations.
%% Of note is the shared affiliation of the first two authors, and the
%% "authornote" and "authornotemark" commands
%% used to denote shared contribution to the research.
\author{Jia Wang}
\affiliation{%
  \institution{Dalian University of Technology}
  \city{Dalian}
  \country{China}
}
\email{jiawang0704@outlook.com}

\author{Jiaming Cai}
\affiliation{%
  \institution{Dalian University of Technology}
  \city{Dalian}
  \country{China}
}
\email{2961885858@mail.dlut.edu.cn}

\author{Zunying Hu}
\affiliation{%
  \institution{Beijing Children's Hospital, Capital Medical University}
  \city{Beijing}
  \country{China}
}
\email{huzy0628@163.com}

\author{Zhanjie Wu}
\affiliation{%
 \institution{\shortstack{Chongqing University of\\Posts and Telecommunications}}
 \city{Chongqing}
 \country{China}
 }
 \email{3195863842@qq.com}

\author{Jinyuan Liu}
\affiliation{%
 \institution{Dalian University of Technology}
 \city{Dalian}
 \country{China}}
  \email{atlantis918@hotmail.com}
 
\author{Hua Cheng}
\affiliation{%
  \institution{Beijing Children's Hospital, Capital Medical University}
  \city{Beijing}
  \country{China}}
\email{64954966@qq.com}

\author{Yun Peng}
\affiliation{%
  \institution{Beijing Children's Hospital, Capital Medical University}
  \city{Beijing}
  \country{China}}
\email{pengyun@mail.ccmu.edu.cn}

\author{Xin Fan}
\correspondingauthor
\affiliation{%
  \institution{Dalian University of Technology}
  \city{Dalian}
  \country{China}
}
\email{xin.fan@dlut.edu.cn}

%%
%% By default, the full list of authors will be used in the page
%% headers. Often, this list is too long, and will overlap
%% other information printed in the page headers. This command allows
%% the author to define a more concise list
%% of authors' names for this purpose.
\renewcommand{\shortauthors}{Wang et al.}

%%
%% The abstract is a short summary of the work to be presented in the
%% article.
\begin{abstract}
Registration-based Few-shot medical image segmentation (RFMIS) aims to generate pseudo-labels for unlabeled images by warping a labeled image through registration. However, existing methods primarily perform pixel-level optimization and inference in Euclidean space, treating anatomical structures as flat and disjoint. This neglect of inherent hierarchies degrades pseudo-label quality and weakens the discrimination of ambiguous regions, limiting the segmentation performance. To overcome this challenge, we propose a Hyperbolic Hierarchy-aware Aggregative Learning framework for RFMIS, termed H²AL, that enhances both deformation plausibility and anatomical discrimination for dual-task learning. Specifically, we introduce a Hyperbolic Hierarchy-aware Infusion (H2I) module, which leverages the hierarchical modeling capability of hyperbolic space to learn precise hierarchy-aware representations via transformation-guided supervised hyperbolic contrastive learning, and injects such hierarchical priors into Euclidean space through a gated infusion block while preserving semantic richness. Furthermore, we propose an end-to-end joint optimization algorithm by gradient aggregation, where the gradients from the registration and segmentation decoders, embedding semantic and hierarchical cues, are aggregated to update the shared encoder to promote collaborative learning across tasks. Extensive experiments on two anatomical regions, with five experimental settings, demonstrate the effectiveness and efficiency of our method in both registration and segmentation. The code is publicly available at \url{https://github.com/JiamingCai469/H2AL}.
\end{abstract}

%%
%% The code below is generated by the tool at http://dl.acm.org/ccs.cfm.
%% Please copy and paste the code instead of the example below.
%%
\begin{CCSXML}
<ccs2012>
   <concept>
       <concept_id>10010405.10010444.10010447</concept_id>
       <concept_desc>Applied computing~Health care information systems</concept_desc>
       <concept_significance>500</concept_significance>
       </concept>
 </ccs2012>
\end{CCSXML}

\ccsdesc[500]{Applied computing~Health care information systems}
% \ccsdesc[500]{Do Not Use This Code~Generate the Correct Terms for Your Paper}
% \ccsdesc[300]{Do Not Use This Code~Generate the Correct Terms for Your Paper}
% \ccsdesc{Do Not Use This Code~Generate the Correct Terms for Your Paper}
% \ccsdesc[100]{Do Not Use This Code~Generate the Correct Terms for Your Paper}

%%
%% Keywords. The author(s) should pick words that accurately describe
%% the work being presented. Separate the keywords with commas.
\keywords{Registration-based Few-shot Medical Image Segmentation, Hyperbolic Hierarchy-aware Infusion, Gradient Aggregation Learning}
%% A "teaser" image appears between the author and affiliation
%% information and the body of the document, and typically spans the
%% page.
% \begin{teaserfigure}
%   \includegraphics[width=\textwidth]{sampleteaser}
%   \caption{Seattle Mariners at Spring Training, 2010.}
%   \Description{Enjoying the baseball game from the third-base
%   seats. Ichiro Suzuki preparing to bat.}
%   \label{fig:teaser}
% \end{teaserfigure}

% \received{20 February 2007}
% \received[revised]{12 March 2009}
% \received[accepted]{5 June 2009}

%%
%% This command processes the author and affiliation and title
%% information and builds the first part of the formatted document.

\begin{teaserfigure}
\centering
\includegraphics[width=\textwidth]{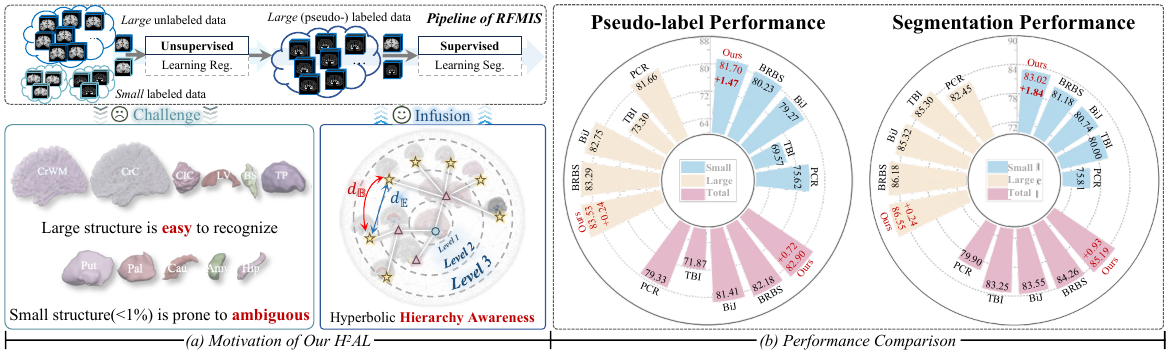}
\vspace{-0.55cm}
\caption{
(a) The upper part shows the pipeline of RFMIS, where unsupervised registration on large unlabeled data produces pseudo-labeled images for supervised segmentation. The bottom row highlights that large anatomical structures are easy to recognize, whereas small structures ($<$1\% volume) are highly ambiguous. The right panel shows our hyperbolic hierarchy-aware modeling, where hyperbolic distance $d_\mathbb B$ between separated regions is much larger than the corresponding Euclidean distance $d_\mathbb E$ ($d_\mathbb B \gg d_\mathbb E$), enlarging inter-class margins and improving discrimination for small structures. (b) Quantitative comparison of pseudo-label (from registration) and segmentation, which reports Dice scores (\%) for large, small, and average structures, with red values indicating the improvement over the second-best results.
}
% \vspace{-0.05cm}
\Description{Overview of the proposed hyperbolic hierarchy-aware FMIS framework and quantitative results.}
\label{fig1}
\end{teaserfigure}

\maketitle

\section{Introduction}

Few-shot medical image segmentation (FMIS) has recently garnered increasing attention as an effective solution for learning from limited labeled data. A prevalent paradigm is registration-based FMIS (RFMIS) which propagates labels via image warping and leverages anatomical correspondence to preserve geometric consistency. By learning unsupervised registration to warp labeled images onto unlabeled ones, it propagates annotations and generates pseudo-labels for training the segmentation network, as shown in Fig.~\ref{fig1}~(a)~(top). \emph{The key to this task lies in the quality of pseudo-labels}, since inaccurate or anatomically inconsistent labels can degrade the segmentation performance, particularly for small anatomical structures.

Existing RFMIS methods~\cite{su2023one, ding2021modeling, lv2023robust, qiu2021rsegnet} mainly focus on strengthening supervision to improve pseudo-label quality. A common strategy is to ensure consistency in appearance and distribution between labeled and unlabeled images. These constraints guide registration to preserve topology and anatomy, and enable the synthesis of style-consistent warped labeled images that augment the scarce labeled set and yield more robust pseudo-labels.
Despite these advancements, most RFMIS methods optimize and infer at the pixel level in Euclidean output embedding spaces, treating anatomical regions as flat and disjoint classes. In practice, medical anatomical categories exhibit an intrinsic coarse-to-fine hierarchy defined by radiologists\footnote{Radiologist-defined coarse-to-fine hierarchy details are in the supplementary.}.
Such hierarchy ignored by existing methods causes confused small or ambiguous anatomical structures (see Fig.~\ref{fig1}~(a)~(down left)), inducing deformation fields that fit local intensities \emph{but violate global hierarchical anatomical consistency}, leading to predicted segmentation results with \emph{weakened anatomical discrimination}.
% Ignoring this structure can confuse small or ambiguous anatomical structures (see Fig.~\ref{fig1}~(a)~(down left)), induces deformation fields that fit local intensities \emph{but violate global anatomical consistency, and ultimately predicts segmentation results with weakened anatomical discrimination}.

To model anatomical hierarchies, recent advanced medical image analysis methods, such as segmentation~\cite{peiris2025co, atigh2022hyperbolic}, registration~\cite{shi2016hyperbolic, zeng20083d}, and image–text alignment~\cite{peng2025understanding, guo2022clipped}, leverage hyperbolic space for representation learning, owing to its exponentially expanding geometry and natural ability to encode hierarchical structures (see Fig.~\ref{fig1}~(a)~(down right)). For example, Hsu et al.~\cite{NEURIPS2021_291d43c6} employed a hyperbolic variational autoencoder and a self-supervised hierarchical loss to learn representations for capturing implicit hierarchical relationships among structures. However, existing hyperbolic methods primarily focus on hyperbolic embedding, \emph{while neglecting the rich semantic information} embedded in Euclidean space, resulting in incomplete modeling of anatomical semantics.  In addition, \emph{no prior methods have investigated its potential in RFMIS}, which necessitates the joint optimization of registration and segmentation and is more challenging than a single-task setting.

Based on the above observations, a key question arises for RFMIS: \textbf{how to effectively leverage both rich semantic information in Euclidean space and anatomical hierarchies in hyperbolic space within a complex dual-task framework.}

% In summary, Euclidean representations excel at capturing rich semantic details but lack hierarchy awareness, while hyperbolic space encodes structural hierarchy but can under-represent subtle spatial details when used in isolation.
% Ideally, hierarchy-aware hyperbolic priors should be injected into the Euclidean feature space while preserving rich anatomical information for RFMIS to achieve more comprehensive feature representations. 

To this end, we propose a Hyperbolic Hierarchy-aware Aggregative Learning framework for RFMIS, named H²AL, that enhances both deformation plausibility and anatomical discrimination for dual-task learning. Building upon a shared encoder and task-specific decoder architecture, Hyperbolic Hierarchy-aware Infusion (H2I) modules explicitly capture anatomical hierarchy from pseudo-label supervision and inject hyperbolic priors into Euclidean features while preserving the semantic richness. 
% Then, the proposed gated infusion block (GIB) aims to inject hyperbolic hierarchical prior information into Euclidean space while preserving the semantic richness.
% Equipped with learned semantic and hierarchical cues, we design a gradient aggregation-based joint optimization strategy that enables efficient one-stage end-to-end training to boost collaborative learning across tasks. 
Equipped with learned semantic and hierarchical cues, we design a gradient aggregation (GA) strategy that enables efficient one-stage end-to-end training by jointly optimizing the shared encoder to boost collaborative learning across tasks.
Overall, H²AL offers a unified framework for anatomically plausible deformation and more discriminative segmentation, especially for small anatomical structures. As shown in Fig.~\ref{fig1}~(b), H²AL leads to \textbf{better pseudo-label quality (left)} and \textbf{superior segmentation performance (right)}. Notably, for small anatomical structures, it improves Dice scores by 1.47\% and 1.84\% for registration and segmentation than SOTA methods respectively.
% Moreover, we design a gradient aggregation-based joint optimization strategy that enables efficient one-stage end-to-end training to boost collaborative learning across tasks. 
Our main contributions are as summarized as:
\begin{itemize}[label=\small$\bullet$,leftmargin=*, labelsep=0.5em, nosep, align=parleft]
\item  We propose H²AL for RFMIS, which enhances both consistent deformation and anatomical representation to increase final segmentation performance.
% \item We introduce a Hyperbolic Hierarchical-aware Infusion  (H2I) module that first explores precise hierarchical representations via hyperbolic contrastive learning, and then fuses these hierarchy-aware features into the Euclidean space while preserving semantic richness.
\item We introduce an H2I module that jointly exploits the hyperbolic hierarchical priors and the Euclidean semantic richness, learning precise and anatomically discriminative representations.
\item We devise a GA-based joint optimization strategy for one-stage end-to-end training, which aggregates task-specific gradients infused with semantic and hierarchical cues to collaboratively optimize the shared encoder across dual tasks.
\item Extensive experiments on brain and cardiac datasets show that H$^2$AL outperforms state-of-the-art RFMIS methods in both segmentation and registration. Comprehensive visualizations further verify the effectiveness of jointly modeling Euclidean semantics and hyperbolic hierarchy in improving the discrimination of small anatomical structures.
\end{itemize}

\begin{figure*}[t]
\centering
\includegraphics[width=\textwidth]{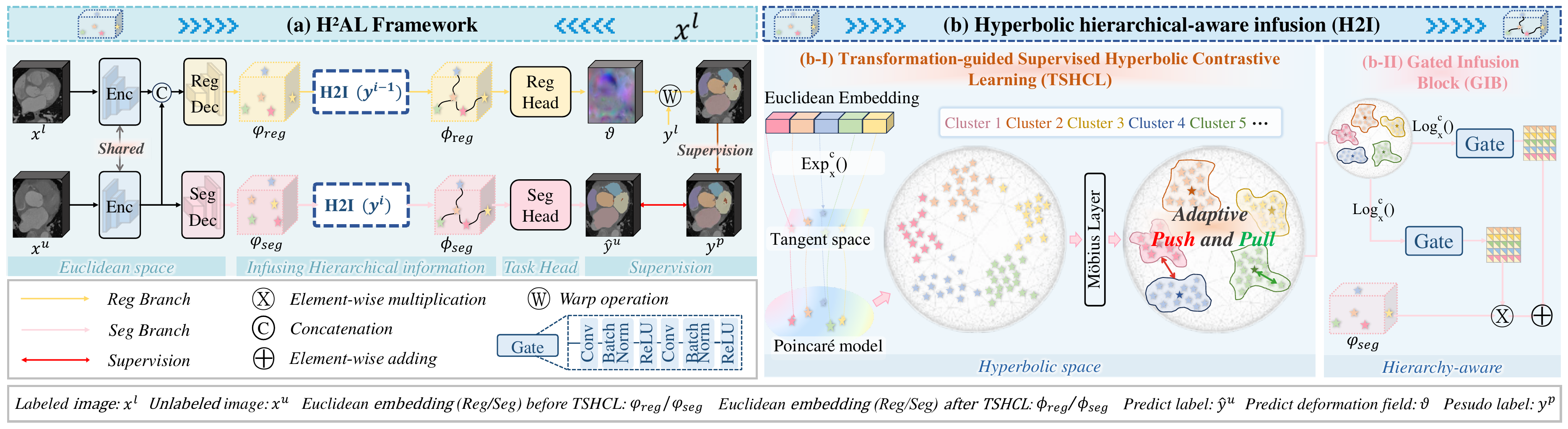}
\vspace{-0.65cm}
\caption{
Overview of the proposed H$^2$AL framework.
(a) The framework jointly performs registration and segmentation with a shared encoder and two task-specific decoders.
(b) H2I consists of two parts: (b-I) TSHCL in hyperbolic space to model class hierarchy with adaptive push–pull, and (b-II) GIB for mapping hyperbolic features back to Euclidean space for final prediction.
}
\Description{Overall architecture of H2AL including shared encoder, dual decoders, hyperbolic contrastive learning, and gated infusion block.}
\vspace{-0.4cm}
\label{fig2}
\end{figure*}

\section{Related Works}

\subsection{Few-shot Medical Image Segmentation}
Few-shot medical image segmentation~\cite{hansen2022anomaly, ouyang2022self, feng2021interactive, cheng2024few, shen2023q} has become an important topic in medical image analysis, which can be divided into two categories: prototype-based and registration-based segmentation. 
Prototype-based methods typically learn class-representative feature embeddings from support images and segment query images through support–query feature matching. Tang et al.~\cite{tang2021recurrent} proposed a prototypical network with a context encoder and recurrent refinement to enhance foreground–background separation. Zhu et al.~\cite{zhu2023few} introduced regional prototypes to reduce intra-class variation and improve support--query matching.
Despite their effectiveness, these methods rely mainly on feature-space similarity and often lack explicit anatomical and spatial modeling, limiting their robustness to large inter-subject variation. 
% 加一个prompt-driven的方法
To address these limitations, registration-based FMIS (RFMIS) has emerged as a more structure-aware alternative by transferring support annotations through learned anatomical alignment. He et al.~\cite{he2022learning} improved registration realism and robustness through knowledge consistency and misalignment regularization, while Fan et al.~\cite{fan2024bi} adopted a bilevel optimization framework jointly learning registration and segmentation with appearance constraints.
Although these methods have achieved promising results, they generally overlook the hierarchical organization of anatomical categories. 
In this work, we focus on registration-based FMIS and offer insights for hyperbolic hierarchical representation learning in broader few-shot settings.
%In this work, we explore hyperbolic representation learning to encode anatomical hierarchies while preserving the Euclidean embedding’s strength in modeling local and discriminative features.

\vspace{-0.2cm}

\subsection{Hyperbolic Hierarchy-Aware Learning}
Recent works have explored hyperbolic hierarchy-aware learning~\cite{liu2025semantic, wei2024poincare, zhang2024hypered, su2023enhancing, wang2021knowledge}, leveraging hyperbolic spaces to model hierarchical structures in data. Dong et al.~\cite{10.1145/3543507.3583376} conducted hierarchical representation and reasoning entirely in hyperbolic space by preserving knowledge graph hierarchies. Pan and Wang~\cite{pan-wang-2021-hyperbolic-hierarchy} developed a hyperbolic embedding method that models hierarchical structures through geometry-aware mapping. However, these methods mainly focus on hyperbolic embedding itself while overlooking the rich semantic information preserved in Euclidean space, which may lead to incomplete representation learning. To address this issue, we inject hierarchy-aware hyperbolic information into Euclidean space to exploit the complementary strengths of both geometries.

\vspace{-0.2cm}

\section{Preliminary: Hyperbolic Geometry}

% Hyperbolic space is a Riemannian manifold with constant negative sectional curvature.
A Riemannian manifold \((\mathcal M,g)\) of dimension \(n\) is a smooth real manifold endowed with a metric \(g\) that assigns each \(x\in\mathcal M\) an inner product \(g_x:T_x\mathcal M\times T_x\mathcal M\to\mathbb R\) on the tangent space \(T_x\mathcal M\) at each point \(x\in\mathbb M\). 
Hyperbolic space is a Riemannian manifold with constant negative curvature, exhibiting exponential volume growth and a tree-like geometry that naturally captures hierarchical relationships, which is beneficial for complex medical data.
% Among the three constant curvature model spaces, \emph{i.e.}, Euclidean \(\mathbb R^n\) (zero), spherical \(\mathbb S^n\) (positive), and hyperbolic (negative), hyperbolic space exhibits exponential volume growth and a tree-like geometry that naturally represents hierarchical relationships, which is advantageous for complex medical data.

\noindent\textbf{Poincar\'e ball model.}
Following the previous work ~\cite{franco2024hyperbolic, hindel2024taxonomy, peiris2025co}, we adopt an open \(n\)-dimensional Poincar\'e ball defined as ($\mathbb B_c^n, g^\mathbb B$), with manifold $\mathbb B_c^n=\bigl\{\ c\|x\|^2<1\,\bigr\}$, where radius is \(1/\sqrt c\). Its Riemannian metric is conformal to the Euclidean metric $g_x^{\mathbb E}$ as
\begin{equation}
  g^{\mathbb B}_x=\lambda_{c,x}^{\,2}\,g^{\mathbb E}_x=\frac{4}{(1-c\|x\|^2)^2}g^{\mathbb E}_x.
  \label{eq:metric}
\end{equation}
Since standard Euclidean vector operations are not directly valid in hyperbolic space, we use M\"obius addition \((\oplus_c)\)  as
% Since traditional vector operation (\emph{e.g.,} addition and scaling) in Euclidean space can not directly apply in curved spaces, the gyrovector space is introduced to define as M\"obius operations \((\oplus_c)\) as
\begin{equation}
  x\oplus_c y
  =\frac{(1+2c\langle x,y\rangle+c\|y\|^2)\,x+(1-c\|x\|^2)\,y}
  {1+2c\langle x,y\rangle+c^2\|x\|^2\|y\|^2},
  \label{eq:mobius}
\end{equation}
\noindent
with analogous M\"obius scalar and matrix multiplications. The hyperbolic distance is
\begin{equation}
  d_{\mathbb B}(x,y)
  =\frac{2}{\sqrt c}\,
  \tanh^{-1}\!\Bigl(\sqrt c\,\|(-x)\oplus_c y\|\Bigr).
  \label{eq:distance}
\end{equation}

\noindent\textbf{Exponential and logarithmic maps.}
The tangent space $\mathcal{T}_x \mathbb{B}_c^n$ at a point $x \in \mathbb{B}_c^n$ is a Euclidean space composed of vectors, aiming to perform optimization and representation learning in hyperbolic space. Specifically, the mapping between the tangent space $\mathcal{T}_x \mathbb{B}_c^n$ and the Poincaré ball model $\mathbb{B}_c^n$ is established by exponential ($\mathcal{T}_x \mathbb{B}_c^n \rightarrow \mathbb{B}_c^n$) and its inverse, the logarithmic ($\mathbb{B}_c^n \rightarrow \mathcal{T}_x \mathbb{B}_c^n$). For any two points $x, y \in \mathbb{B}_c^n$ and a tangent vector $v \in \mathcal{T}_x \mathbb{B}_c^n$, where $v \neq 0$ and $y \neq x$, the map function is given by
\begin{equation}
  \exp_x^{\mathbb B_c}(v)
  =x\oplus_c\!\left(
  \tanh\!\left(\frac{\sqrt c\,\lambda_{c,x}\,\|v\|}{2}\right)
  \frac{v}{\sqrt c\,\|v\|}
  \right),
  \label{eq:exp}
\end{equation}
\begin{equation}
  \log_x^{\mathbb B_c}(y)
  =\frac{2}{\sqrt c\,\lambda_{c,x}}\,
  \tanh^{-1}\!\Bigl(\sqrt c\,\|(-x)\oplus_c y\|\Bigr)\,
  \frac{(-x)\oplus_c y}{\|(-x)\oplus_c y\|}.
  \label{eq:log}
\end{equation}

\section{Methodology}
\subsection{Framework Overview}

Registration-based few-shot medical image segmentation (RFMIS) exploits registration (reg.) to generate pseudo-labels, thereby facilitating segmentation (seg.) under limited supervision. In the few-shot setting, the training data consists of a small labeled subset, denoted as $\mathcal{D}_l = \{(x_i^l, y_i^l)\}_{i=1}^{N}$, and a larger unlabeled subset, denoted as $\mathcal{D}_u = \{x_j^u\}_{j=1}^{M}$, where $N \ll M$. Here, each 3D medical volume is $x^l \in \mathbb{R}^{D \times H \times W}$, and the corresponding ground truth segmentation is $y^l \in \{0, 1, ..., C{-}1\}^{D \times H \times W}$. For brevity, the sample indices $i$ and $j$ are omitted in the following descriptions.

To this end, we propose a Hyperbolic Hierarchy-aware Aggregative Learning framework, termed H²AL, which consists of a shared encoder and two task-specific decoders, H2I modules, and dedicated task heads for registration and segmentation, respectively. As shown in Fig.~\ref{fig2}, both labeled and unlabeled volumes are first fed into the shared encoder to extract their feature representations. The unlabeled features are then concatenated with the labeled branch features and passed to the registration decoder to produce the deformation field embeddings $\phi_{reg}$. Meanwhile, the unlabeled features are forwarded through the segmentation decoder to generate the segmentation embeddings $\phi_{seg}$.
\emph{To effectively leverage anatomical prior knowledge}, the deformation embedding is further fed into the H2I module to inject hyperbolic hierarchical information into the Euclidean feature space. Specifically, the H2I module consists of a Transformation-guided Supervised Hyperbolic Contrastive Learning (TSHCL) strategy and a Gated Infusion Block (GIB). The resulting hierarchy-aware features are finally delivered to the registration head to predict the final deformation field $\vartheta$
. The segmentation branch follows a similar process to generate the predicted label $\hat{y}^u$. It is worth noting that for the registration branch, the TSHCL module is supervised using the previous-iteration pseudo-label (\emph{i.e.,} reg.-warped label), while for the segmentation branch, it uses the current pseudo-label. Finally, aiming at a more robust and generalizable encoder, we propose a \emph{gradient aggregation} strategy to effectively integrate Euclidean and hyperbolic cues, yielding stable optimization and stronger representations.

\vspace{-0.5cm}

\subsection{Hyperbolic Hierarchy-aware Infusion}
\label{H2C}
Given the Euclidean embedding $\phi$, we propose the hyperbolic hierarchy-aware Infusion (H2I) module to inject hierarchical information from hyperbolic space into Euclidean feature maps. The H2I module consists of two key components: \emph{(1) Transformation-guided supervised hyperbolic contrastive learning (TSHCL)} strategy, which exploits hierarchy-aware representations in hyperbolic space. \emph{(2) Gated infusion block (GIB)}, which injects these representations into the Euclidean feature stream to learn richer feature representations.

\noindent\textbf{TSHCL strategy.}
To model the implicit hierarchical relationships in medical data, we map Euclidean embeddings into hyperbolic space using the Poincar\'e ball model. Leveraging the natural ability of hyperbolic geometry to model tree-like structures, this yields more discriminative and hierarchy-aware representation learning. Such hierarchical embeddings are particularly beneficial for improving downstream registration and segmentation performance.

To this end, we propose a Transformation-guided Supervised Hyperbolic Contrastive Learning (TSHCL) strategy, which adaptively pushes and pulls class embeddings based on their hyperbolic distances with pseudo-labels from registration. \emph{Unlike previous methods that solely rely on the hyperbolic geometry for implicit representation learning, we incorporate multi-label supervision to explicitly guide the formation of a structured embedding space in the Poincar\'e ball}. Specifically, given feature embeddings \( \{z_i\}_{i=1}^{N} \subset \mathbb{B}_c^n \) and pseudo-labels \( \{\hat{y}_i^p\}_{i=1}^{N} \) obtained from the registration branch, we first define a binary mask that distinguishes positive and negative pairs as
\begingroup
\setlength{\abovedisplayskip}{4pt}
\setlength{\belowdisplayskip}{4pt}
\setlength{\abovedisplayshortskip}{2pt}
\setlength{\belowdisplayshortskip}{2pt}
\begin{equation}
\mathbb{I}_{i,j} =
\begin{cases}
1, & \text{if } \hat{y}_i = \hat{y}_j \text{ and } i \ne j,\\
0, & \text{otherwise.}
\end{cases}
\label{eq:mask}
\end{equation}
\endgroup
We compute pairwise similarity between hyperbolic embeddings using the geodesic distance in Eq.~\ref{eq:distance}. Accordingly, we design distance-aware contrastive weights to regulate attractive and repulsive forces: a pull weight for positive pairs that increases with distance to promote compactness, and a push weight for negative pairs that decays with distance to avoid excessive repulsion between unrelated samples. The corresponding weight functions are defined as
\begingroup
\setlength{\abovedisplayskip}{4pt}
\setlength{\belowdisplayskip}{4pt}
\setlength{\abovedisplayshortskip}{2pt}
\setlength{\belowdisplayshortskip}{2pt}
\begin{gather}
w^{{pull}}_{i,j} = \left(1 - e^{-\gamma \cdot d_{\mathbb{B}}(z_i, z_j)}\right), \label{eq:weights}\\
w^{{push}}_{i,j} = e^{-\delta \cdot d_{\mathbb{B}}(z_i, z_j)}.
\end{gather}
\endgroup
This design ensures stronger attraction between dissimilar but true positive samples, and softer repulsion between negatives that are already distant.
Finally, the TSHCL loss is defined as
\begingroup
\setlength{\abovedisplayskip}{4pt}
\setlength{\belowdisplayskip}{4pt}
\setlength{\abovedisplayshortskip}{2pt}
\setlength{\belowdisplayshortskip}{2pt}
\begin{equation}
\mathcal{L}_{\text{TSHCL}} = 
\sum_{i\in D} \log \frac{
\sum\limits_{j\in D} \mathbb{I}_{i,j}  \exp \left( w^{\text{pull}}_{i,j}/{\tau} \right)
}{
\sum\limits_{j\in D} \exp \left( \left(w^{\text{pull}}_{i,j} \mathbb{I}_{i,j} + w^{\text{push}}_{i,j}  (1 - \mathbb{I}_{i,j})\right)/{\tau} \right)
},
\label{eq:loss}
\end{equation}
\endgroup
where $\tau$ denotes the temperature parameter. With the proposed TSHCL strategy, we further adopt M\"{o}bius layer defined in Eq.~\ref{eq:mobius} to refine hyperbolic embeddings. Thus, leveraging the geometric properties of hyperbolic space, we can generate embeddings that are structurally aligned with the underlying hierarchy.

\begin{figure}[t]
    \centering
    \includegraphics[width=\linewidth]{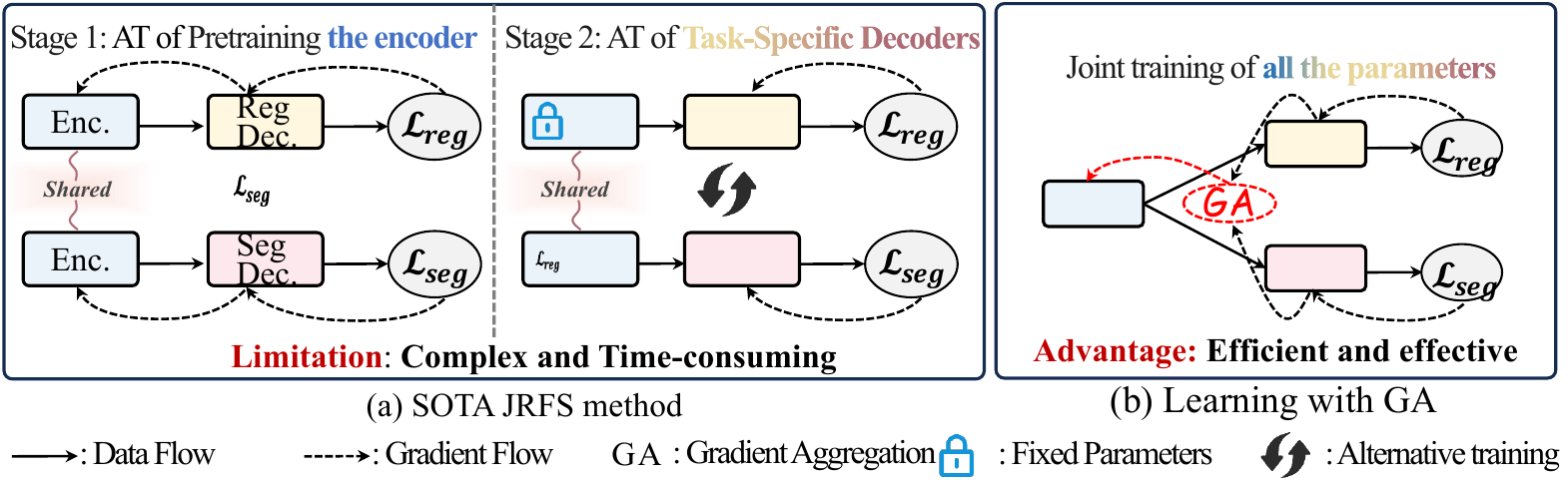}
    \vspace{-0.65cm}
    \caption{Comparison between the state-of-the-art RFMIS method and our proposed learning strategy with gradient aggregation (GA).}
    \Description{Visual comparison between RFMIS and the proposed gradient aggregation training strategy.}
    \vspace{-0.5cm}
    \label{fig:fig3}
\end{figure}

\begin{table*}[!t]
\centering
\caption{Quantitative results of segmentation methods on the Brain and the Cardiac datasets. The numbers in parentheses after each dataset name indicate the modality of the dataset.}
\vspace{-0.2cm}
\label{tab1}
\renewcommand{\arraystretch}{1.10}
\resizebox{\linewidth}{!}{
\begin{tabular}{c||c|cccccc|cccc}
\toprule[0.3mm]
\multicolumn{12}{c}{
  \coloredcircle{myblue}{F} : Fully supervised \quad
  \coloredcircle{mygreen}{A} : Atlas-based \quad
  \coloredcircle{mypurple}{F} : Few-shot \quad
  \textcolor{lightpink}{\rule{10pt}{7pt}} : Best performance \quad
  \textcolor{lightblue}{\rule{10pt}{7pt}} : Second best performance
} \\
\bottomrule
\toprule
\multicolumn{1}{c||}{\multirow{3}{*}{Methods}} &
\multirow{3}{*}{Type} &
\multicolumn{6}{c|}{Brain(T1 MRI)} & \multicolumn{4}{c}{Cardiac(CT)} \\ \cmidrule{3-12} 
\multicolumn{1}{c||}{} & &
\multicolumn{2}{c|}{Atlas\textsubscript{±std}} &
\multicolumn{2}{c|}{1-shot\textsubscript{±std}} &
\multicolumn{2}{c|}{5-shot\textsubscript{±std}} &
\multicolumn{2}{c|}{1-shot\textsubscript{±std}} &
\multicolumn{2}{c}{5-shot\textsubscript{±std}} \\
\multicolumn{1}{c||}{} & &
Dice\%$\uparrow$ & \multicolumn{1}{c|}{HD\textsubscript{95}mm$\downarrow$} &
Dice\%$\uparrow$ & \multicolumn{1}{c|}{HD\textsubscript{95}mm$\downarrow$} &
Dice\%$\uparrow$ & \multicolumn{1}{c|}{HD\textsubscript{95}mm$\downarrow$} &
Dice\%$\uparrow$ & \multicolumn{1}{c|}{HD\textsubscript{95}mm$\downarrow$} &
Dice\%$\uparrow$ & \multicolumn{1}{c}{HD\textsubscript{95}mm$\downarrow$} \\ 
\midrule
UNet~\cite{ronneberger2015u} & \coloredcircle{myblue}{F} & 46.36 \textsubscript{± 19.87} & 19.83 \textsubscript{± 8.37} & 53.41 \textsubscript{± 16.83} & 13.75 \textsubscript{± 5.79} & 52.18 \textsubscript{± 18.73} & 9.90 \textsubscript{± 7.08} & 57.59 \textsubscript{± 16.38} & 33.70 \textsubscript{± 7.35} & 85.96 \textsubscript{± 9.57} & 9.38 \textsubscript{± 7.96} \\
MASSL~\cite{chen2019multi} & \coloredcircle{myblue}{F} & 58.01 \textsubscript{± 2.16} & 15.63 \textsubscript{± 7.11} & 61.73 \textsubscript{± 2.28} & 10.47 \textsubscript{± 5.67} & 60.15 \textsubscript{± 0.05} & 8.82 \textsubscript{± 4.38} & 61.23 \textsubscript{± 5.86} & 22.94 \textsubscript{± 6.71} & 65.49 \textsubscript{± 5.03} & 15.53 \textsubscript{± 3.07} \\
Co-manifold~\cite{peiris2025co} & \coloredcircle{myblue}{F} & 64.41 \textsubscript{± 6.72} & 12.07 \textsubscript{± 1.70} & 66.28 \textsubscript{± 6.64} & 11.52 \textsubscript{± 1.01} & 67.94 \textsubscript{± 5.92} & 8.03 \textsubscript{± 0.77} & 65.36 \textsubscript{± 6.88} & 19.58 \textsubscript{± 1.92} & 69.55 \textsubscript{± 7.04} & 13.42 \textsubscript{± 1.39} \\
\midrule
DeepAtlas~\cite{10.1007/978-3-030-32245-8_47} & \coloredcircle{mygreen}{A} & 70.15 \textsubscript{± 8.20} & 2.90 \textsubscript{± 0.67} & 67.87 \textsubscript{± 5.45} & 2.74 \textsubscript{± 0.97} & 70.36 \textsubscript{± 9.86} & 3.14 \textsubscript{± 0.92} & 73.43 \textsubscript{± 5.59} & 10.21 \textsubscript{± 3.21} & 82.29 \textsubscript{± 4.42} & 10.59 \textsubscript{± 4.45} \\
DataAug~\cite{cubuk2019autoaugment} & \coloredcircle{mygreen}{A} & 73.52 \textsubscript{± 3.51} & 3.57 \textsubscript{± 0.87} & 76.39 \textsubscript{± 2.55} & 2.32 \textsubscript{± 0.35} & 75.69 \textsubscript{± 2.65} & 2.65 \textsubscript{± 0.41} & 82.26 \textsubscript{± 7.21} & 10.16 \textsubscript{± 4.08} & 83.01 \textsubscript{± 5.32} & 9.32 \textsubscript{± 2.15} \\
TBIOneShot~\cite{zhao2023one} & \coloredcircle{mygreen}{A} & 80.60 \textsubscript{± 6.82} & 2.10 \textsubscript{± 0.31} & 83.51 \textsubscript{± 5.97} & 1.55 \textsubscript{± 0.15} & 84.30 \textsubscript{± 5.21} & 1.60 \textsubscript{± 0.25} & 84.30 \textsubscript{± 10.42} & 10.07 \textsubscript{± 4.59} & 88.93 \textsubscript{± 7.51} & 6.66 \textsubscript{± 3.86}\\
\midrule
BRBS~\cite{he2022learning} & \coloredcircle{mypurple}{F} & 82.86 \textsubscript{± 1.45} & 1.75 \textsubscript{± 0.47} & \cellcolor{lightblue}{84.26 \textsubscript{± 1.48}} & \cellcolor{lightblue}{1.49 \textsubscript{± 0.16}} & \cellcolor{lightblue}{85.03 \textsubscript{± 1.78}} & \cellcolor{lightblue}{1.55 \textsubscript{± 0.21}} & 88.31 \textsubscript{± 2.87} & 4.76 \textsubscript{± 1.94} & \cellcolor{lightpink}{90.38 \textsubscript{± 3.76}} & \cellcolor{lightpink}{3.00 \textsubscript{± 1.34}} \\
PC-Reg-RT~\cite{he2021few} & \coloredcircle{mypurple}{F} & 78.30 \textsubscript{± 3.01} & 2.15 \textsubscript{± 0.33} & 79.90 \textsubscript{± 2.12} & 1.82 \textsubscript{± 0.25} & 79.14 \textsubscript{± 3.43} & 1.92 \textsubscript{± 0.36} & 82.16 \textsubscript{± 8.59} & 13.28 \textsubscript{± 9.10} & 89.42 \textsubscript{± 3.01} & 4.01 \textsubscript{± 2.23} \\
Bi-JROS~\cite{fan2024bi} & \coloredcircle{mypurple}{F} & \cellcolor{lightblue}{83.04 \textsubscript{± 1.08}} & \cellcolor{lightblue}{1.74 \textsubscript{± 0.16}} & 83.55 \textsubscript{± 1.41} & 1.50 \textsubscript{± 0.16} & 84.75 \textsubscript{± 1.86} & 1.74 \textsubscript{± 0.28} & \cellcolor{lightblue}{88.35 \textsubscript{± 3.11}} & \cellcolor{lightblue}{4.41 \textsubscript{± 1.59}} & 89.81 \textsubscript{± 3.10} & 4.38 \textsubscript{± 1.09} \\
\midrule
Ours & \coloredcircle{mypurple}{F} & \cellcolor{lightpink}{84.17 \textsubscript{± 1.09}} & \cellcolor{lightpink}{1.70 \textsubscript{± 0.17}} & \cellcolor{lightpink}{85.19 \textsubscript{± 1.23}} & \cellcolor{lightpink}{1.41 \textsubscript{± 0.14}} & \cellcolor{lightpink}{85.68 \textsubscript{± 1.52}} & \cellcolor{lightpink}{1.32 \textsubscript{± 0.15}} & \cellcolor{lightpink}{89.24 \textsubscript{± 2.11}} & \cellcolor{lightpink}{4.15 \textsubscript{± 1.47}} & \cellcolor{lightblue}{90.24 \textsubscript{± 2.01}} & \cellcolor{lightblue}{3.56 \textsubscript{± 1.30}} \\
\bottomrule[0.3mm]
\end{tabular}%
}
\end{table*}

\begin{figure*}[t]
    \centering
    \includegraphics[width=\linewidth]{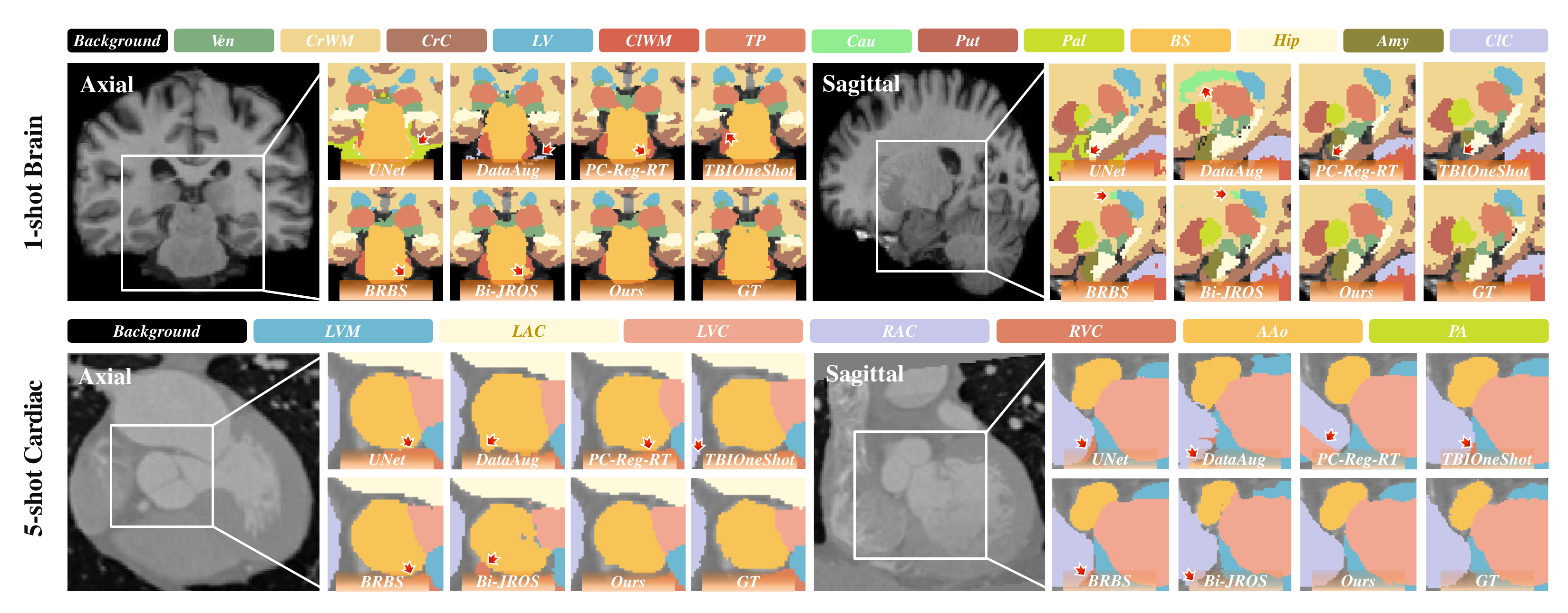}
    \vspace{-0.75cm}
    \caption{Qualitative segmentation results on the Brain and Cardiac datasets. Different colors indicate different anatomical structures, whose names are shown above the panels. Red arrows highlight erroneous predictions.}
    \vspace{-0.1cm}
    \Description{Qualitative comparisons across methods on brain and cardiac images, highlighting segmentation differences; red arrows mark failure cases.}
    \label{fig:seg}
\end{figure*}

\noindent\textbf{GIB.}
Leveraging the rich hierarchical representations encoded in hyperbolic space, we propose the Gated Infusion Block (GIB) to inject such structural knowledge into Euclidean features for downstream registration and segmentation tasks. Specifically, GIB first projects the hierarchy-aware hyperbolic embeddings into Euclidean space, as defined in Eq.~\ref{eq:log}, to ensure geometric compatibility. 
% This design allows the model to preserve the semantic richness of Euclidean representations while effectively incorporating the hierarchy-aware priors encoded in hyperbolic geometry.

Then, the projected embeddings are passed through a gated attention mechanism to generate hierarchy-informed attention maps, which are subsequently used to modulate the Euclidean features via an affine transformation as
\begin{equation}
   \phi = \varphi \times Gate_1(\log^{\mathbb{B}_c}_x(\mathbf{\theta})) + Gate_2(\log^{\mathbb{B}_c}_x(\mathbf{\theta})),
\end{equation}
where $\varphi$, $\theta$, and $\phi$ denote the Euclidean embedding, hyperbolic embedding, and the Euclidean embedding with hierarchical information.
The proposed GIB employs a simple yet effective design that takes advantage of the rich semantic representations from the Euclidean space while incorporating the hierarchical priors embedded in hyperbolic geometry. To better adapt to different tasks, we instantiate two task-specific GIBs for registration and segmentation, respectively, allowing each task to selectively exploit the infused hierarchical cues. This fusion yields task-specific embeddings that are simultaneously semantically discriminative and hierarchically structured, leading to improved registration accuracy and segmentation robustness.

\subsection{Network Architecture and Optimization}

Existing RFMIS methods adopt the shared-encoder and task-specific framework~\cite{zhao2021deep,fan2024bi}, proving to be an effective framework. Thus, we also adopt this framework as the basis of our method due to its outstanding performance.
However, these methods often employ a two-stage training strategy that the shared encoder is pretrained and then followed by alternating optimization of task-specific decoders, resulting in time-consuming and suboptimal task interaction modeling as shown in Fig.~\ref{fig:fig3}~(a). \emph{While the proposed H2I effectively enriches the feature representations with structural priors, a suitable training strategy is required to fully exploit such enriched features across tasks.}

We propose a one-stage, end-to-end training strategy, termed \textbf{learning with Gradient Aggregation (GA)}, that enables efficient and consistent multi-task learning as shown in Fig.~\ref{fig:fig3}~(b), and the pseudo-code is shown in Algorithm 1. Specifically, we first perform task-specific decoder updates during backpropagation as
\begin{equation}
\theta_{rd} \leftarrow \theta_{rd} - \nabla_{\theta_{rd}} \mathcal{L}_{\text{reg}},
\end{equation}
\begin{equation}
\theta_{sd} \leftarrow \theta_{sd} - \nabla_{\theta_{sd}} \mathcal{L}_{\text{seg}},
\end{equation}
where $\theta_{rd}$ and $\theta_{sd}$ denote the parameters of registration and segmentation decoders, and $\mathcal{L}_{reg}$ and $\mathcal{L}_{seg}$ denote the loss function of registration and segmentation task. After obtaining the gradients from the task-specific decoders embedding semantic and hierarchical cues, we aggregate them to update the shared encoder in a unified manner as
\begin{equation}
\theta_{se} \leftarrow \theta_{se} - 
\big( 
\nabla_{\theta_{se}} \mathcal{L}_{reg} + \nabla_{\theta_{se}} \mathcal{L}_{seg}
\big),
\end{equation}
where $\theta_{se}$ denotes the parameter of the shared encoder. By combining gradients before applying updates to $\theta{se}$, we facilitate synergistic representation learning, encouraging the encoder to learn features that are beneficial to both registration and segmentation tasks. This tightly coupled training regime allows the model to better capture inter-task dependencies and avoid task dominance.

\noindent\textbf{Reg Loss.}
We use a smoothness loss $\mathcal{L}_{\text{smo}}$ to enforce deformation smoothness, and adopt $\mathcal{L}_{sim}$ to measure the similarity between the warped image and target image.
The total registration loss with the proposed TSHCL strategy is then a weighted sum of these components as
\begin{equation}
\mathcal{L}_{\text{reg}} = \lambda_{smo} \mathcal{L}_{\text{smo}} + \lambda_{sim} \mathcal{L}_{\text{sim}} + \lambda_{TSHCL}^{reg}\mathcal{L}_{TSHCL}^{reg},
\end{equation}
where $\lambda_{smo}$, $\lambda_{sim}$, $\lambda_{TSHCL}^{reg}$ are the hyper-parameters to balance the trade-off of the three components.

\noindent\textbf{Seg loss.} We take a multi-class Dice coefficient loss as the segmentation metric with the proposed TSHCL as
\begin{equation}
\mathcal{L}_{\text{seg}} = \lambda_{dice}\mathcal{L}_{\text{dice}}+\lambda_{TSHCL}^{seg}\mathcal{L}_{TSHCL},
\end{equation}
% where $\lambda_{dice}$ are $\lambda_{HCL}^{seg}$ are the hyperparameters that balance the two components.
where $\lambda_{dice}$ and $\lambda_{TSHCL}^{seg}$ are the hyperparameters used to balance the relative contributions of the two components.

\begin{table}[t]
\centering
\caption{Quantitative comparison of registration performance among different methods on the Brain dataset under the 5-shot setting. The symbols \captioncircle{mygreen}{A} and \captioncircle{mypurple}{F} denote atlas-based and few-shot methods, respectively.}
\vspace{-0.15cm}
\label{tab:reg_results_5shot}
\resizebox{\linewidth}{!}{%
\begin{tabular}{>{\centering\arraybackslash}p{2.4cm}|c|>{\centering\arraybackslash}p{1.8cm}>{\centering\arraybackslash}p{1.8cm}>{\centering\arraybackslash}p{1.8cm}}
\toprule
\multicolumn{5}{c}{
  \coloredcircle{myblue}{R} : Registration-only \,
  \textcolor{lightpink}{\rule{10pt}{7pt}} : Best performance \,
  \textcolor{lightblue}{\rule{10pt}{7pt}} : Second best performance
} \\
\bottomrule
\toprule
\textbf{Methods} & \textbf{Type} & \textbf{Dice\%$\uparrow$} & \textbf{NCC$\uparrow$} & \textbf{$|J_\phi| \le 0 \ \% \downarrow$} \\
\midrule
Initial & -- & 63.04 \textsubscript{± 7.13} & 0.151 \textsubscript{± 0.012} & -- \\
\midrule
SyN~\cite{avants2011reproducible} & \coloredcircle{myblue}{R} & 77.42 \textsubscript{± 3.54} & 0.269 \textsubscript{± 0.014} & 0.00 \textsubscript{± 0.00} \\
VoxelMorph~\cite{balakrishnan2019voxelmorph} & \coloredcircle{myblue}{R} & 71.13 \textsubscript{± 8.92} & 0.258 \textsubscript{± 0.027} & 0.42 \textsubscript{± 0.35} \\
\midrule
DeepAtlas~\cite{10.1007/978-3-030-32245-8_47} & \coloredcircle{mygreen}{A} & 64.12 \textsubscript{± 14.08} & 0.154 \textsubscript{± 0.011} & 3.12 \textsubscript{± 0.56} \\
DataAug~\cite{cubuk2019autoaugment} & \coloredcircle{mygreen}{A} & 78.27 \textsubscript{± 3.56} & 0.365 \textsubscript{± 0.010} & 0.52 \textsubscript{± 0.09} \\
TBIOneShot~\cite{zhao2023one} & \coloredcircle{mygreen}{A} & 74.02 \textsubscript{± 2.83} & 0.297 \textsubscript{± 0.008} & 0.00 \textsubscript{± 0.00} \\
\midrule
BRBS~\cite{he2022learning} & \coloredcircle{mypurple}{F} & \cellcolor{lightblue}82.53 \textsubscript{± 1.81} & 0.335 \textsubscript{± 0.009} & 2.86 \textsubscript{± 0.33} \\
PC-Reg-RT~\cite{he2021few} & \coloredcircle{mypurple}{F} & 76.45 \textsubscript{± 4.43} & 0.289 \textsubscript{± 0.014} & 1.59 \textsubscript{± 0.22} \\
Bi-JROS~\cite{fan2024bi} & \coloredcircle{mypurple}{F} & 81.47 \textsubscript{± 2.34} & \cellcolor{lightblue}0.377 \textsubscript{± 0.011} & 0.15 \textsubscript{± 0.07} \\
\midrule
Ours & \coloredcircle{mypurple}{F} & \cellcolor{lightpink}83.36 \textsubscript{± 1.64} & \cellcolor{lightpink}0.386 \textsubscript{± 0.011} & 0.36 \textsubscript{± 0.09} \\
\bottomrule
\end{tabular}%
}
\end{table}

\begin{figure}[t]
    \centering
    \includegraphics[width=\linewidth]{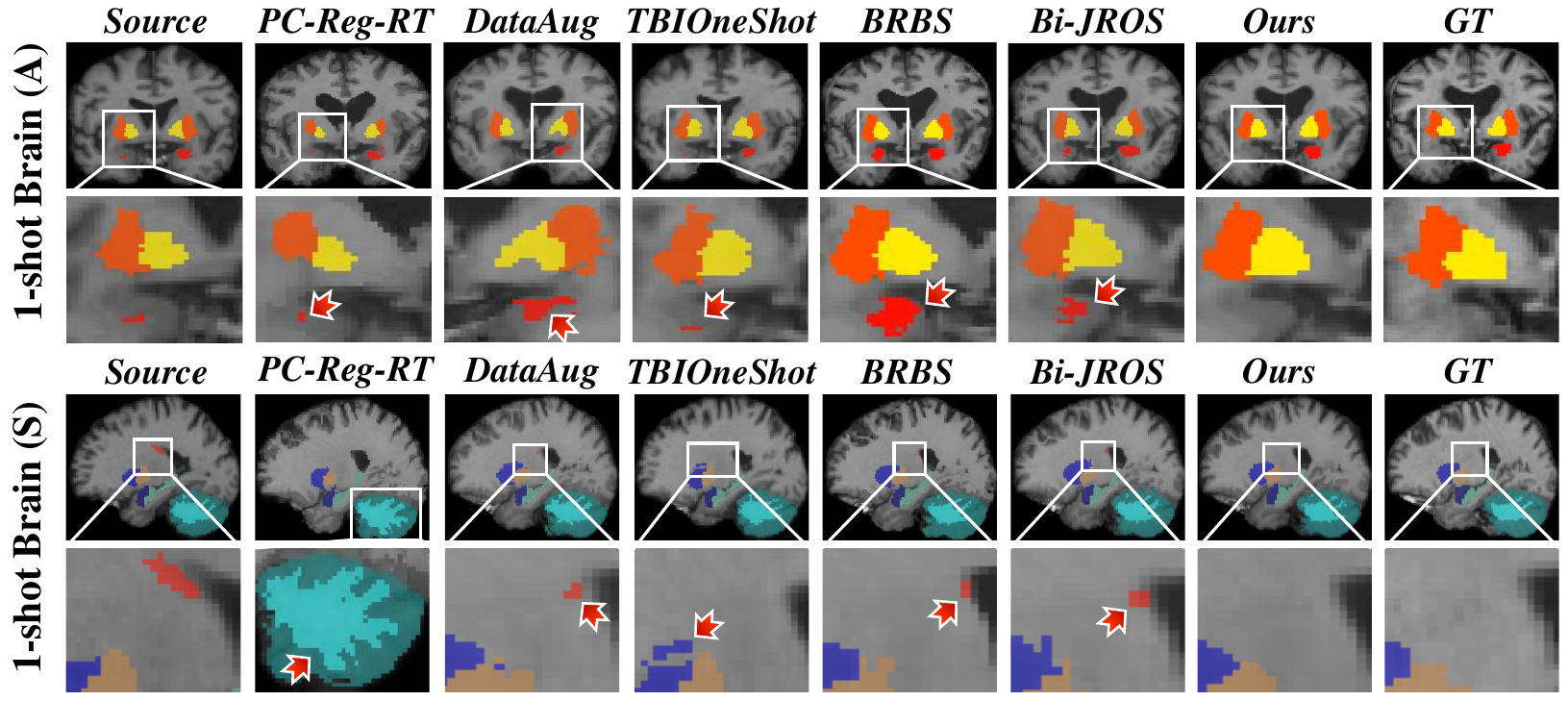}
    \vspace{-0.55cm}
    \caption{Qualitative registration results on the Brain dataset. “A” and “S” denote axial and sagittal views, respectively. Red arrows highlight misaligned regions.}
    \vspace{-0.3cm}
    \Description{Qualitative registration results shown in axial and sagittal views for multiple methods; red arrows highlight mismatches.}
    \label{fig:reg}
\end{figure}

\section{Experiments}
We evaluate H$^2$AL from four perspectives: overall superiority over state-of-the-art methods, behavior under challenging cases, mechanism of performance gains, and reliability of the proposed design. We first describe the experimental settings, datasets, implementation details, and evaluation metrics.
% \subsection{Datasets}

\noindent\textbf{Experimental Settings and Datasets.} We evaluate the proposed method under three settings: Atlas-based, 1-shot, and 5-shot, on the Brain and Cardiac datasets. The Brain dataset is used in all three settings, while the Cardiac dataset is evaluated under the 1-shot and 5-shot settings. In the few-shot setting, instead of using a predefined atlas dataset, we randomly select one or five labeled images from the same dataset as the atlas set. All images and labels are resampled to a uniform size after intensity normalization and label mapping, respectively.
The \textbf{Brain} dataset comprises 410 3D T1-weighted MRI scans from four public datasets: OASIS~\cite{marcus2010open}, PPMI~\cite{marek2011parkinson}, ADNI~\cite{mueller2005ways}, and ABIDE~\cite{di2014autism}. 
These data span a wide range of ages and imaging conditions, providing diverse anatomical variability for model evaluation. 
Following standard practice, 330 volumes are used for training and 80 for testing. 
All scans are resampled to $128 \times 128 \times 128$.
The \textbf{Cardiac} dataset comprises 125 3D CT scans from MM-WHS~\cite{zhuang2019evaluation}, ASOCA~\cite{gharleghi2022asoca}, and CAT08~\cite{schaap2009standardized}, following BRBS~\cite{he2022learning}.
The data include different cardiac phases and anatomical variations, supporting evaluation under realistic inter-subject and inter-site conditions. 
Among them, 100 images are used for training and 25 for testing. 
Each image is resampled to $144 \times 144 \times 128$.

\noindent\textbf{Implementation details.} The proposed method is implemented in PyTorch and trained on an NVIDIA A40 GPU with 48 GB memory. We apply consistent 3D spatial augmentations, including random rotations and isotropic scaling. The registration and segmentation branches are jointly optimized using a single Adam optimizer. The initial learning rate is set to $1\times10^{-4}$, and the models are trained with a batch size of 1.

\noindent\textbf{Metrics.} We adopt three standard metrics for evaluation: Dice, NCC, and HD95. Dice measures the volumetric overlap between prediction and ground truth. NCC evaluates the intensity alignment between the warped source image and the target. HD95 quantifies boundary accuracy using the 95th-percentile Hausdorff distance.

\begin{figure*}[t]
    \centering
    \includegraphics[width=\linewidth]{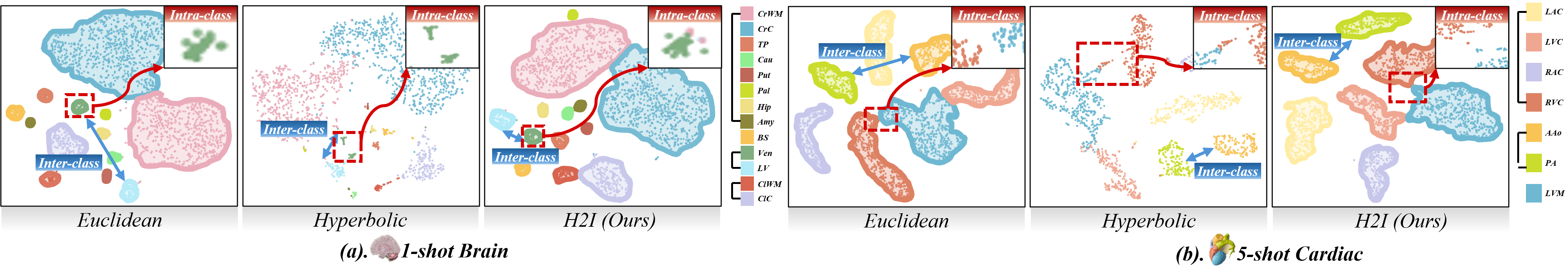}
    \vspace{-0.7cm}
    \caption{
        t-SNE visualizations of feature embeddings learned from the segmentation branch for the Brain (a) and Cardiac (b) datasets. The legend reflects the hierarchical relationships among anatomical structures.
    }
    \vspace{-0.3cm}
    \Description{
        t-SNE plots showing feature distributions for brain and cardiac datasets across different representation spaces. Colors denote anatomical classes; the legend diagram shows hierarchical structure relationships.
    }
    \label{fig:enter-label}
\end{figure*}

\subsection{Comparison with State-of-the-art Methods}

We comprehensively evaluate the segmentation and registration performance of H²AL through both quantitative and qualitative analyses on multiple benchmarks and settings.

\noindent\textbf{Quantitative Comparison.}
For segmentation, we compare H²AL with nine representative methods from three categories, including:
i) fully supervised methods, UNet~\cite{ronneberger2015u}, MASSL~\cite{chen2019multi}, and Co-manifold~\cite{peiris2025co};
ii) atlas-based methods, DeepAtlas~\cite{10.1007/978-3-030-32245-8_47}, DataAug~\cite{cubuk2019autoaugment}, and TBIOneShot~\cite{zhao2023one}, which rely on a single annotated atlas for label propagation via registration;
iii) few-shot methods, BRBS~\cite{he2022learning}, PC-Reg-RT~\cite{he2021few}, and Bi-JROS~\cite{fan2024bi}.

The segmentation results are summarized in Table~\ref{tab1}. Overall, few-shot methods outperform both fully supervised and atlas-based methods, demonstrating their effectiveness in low-label scenarios. Although not specifically designed for the atlas-based setting, H$^2$AL consistently achieves superior performance, demonstrating strong generalization. Attributed to enhanced anatomical discrimination enabled by hierarchy-aware representation learning, H$^2$AL achieves the best performance across all settings on the Brain dataset, significantly outperforming leading few-shot methods such as BRBS and Bi-JROS. On the Cardiac dataset, it ranks first in the 1-shot setting and second in the 5-shot setting, consistently surpassing Bi-JROS.

For registration, we compare our method against eight representative approaches spanning three categories:
i) registration-only methods, SyN~\cite{avants2011reproducible} and VoxelMorph~\cite{balakrishnan2019voxelmorph};
ii) atlas-based methods, DeepAtlas~\cite{10.1007/978-3-030-32245-8_47}, DataAug~\cite{cubuk2019autoaugment}, and TBIOneShot~\cite{zhao2023one};
iii) few-shot methods, BRBS~\cite{he2022learning}, PC-Reg-RT~\cite{he2021few}, and Bi-JROS~\cite{fan2024bi}.

As shown in Table~\ref{tab:reg_results_5shot}, both atlas-based and few-shot methods consistently outperform registration-only approaches, underscoring the importance of label supervision for anatomically consistent deformation. Building on this paradigm, H$^2$AL further leverages hierarchical anatomical information by first aligning coarse parent structures and then refining finer child structures, achieving the highest Dice and NCC scores.
Importantly, these gains are not obtained at the expense of topology preservation. H$^2$AL maintains a low percentage of non-diffeomorphic regions ($|J_\phi| \leq 0$), outperforming most competing methods. This demonstrates a favorable balance between deformation flexibility and topology preservation, leading to more reliable and anatomically consistent registration.

\begin{figure}[!t]
    \centering
    \includegraphics[width=\linewidth]{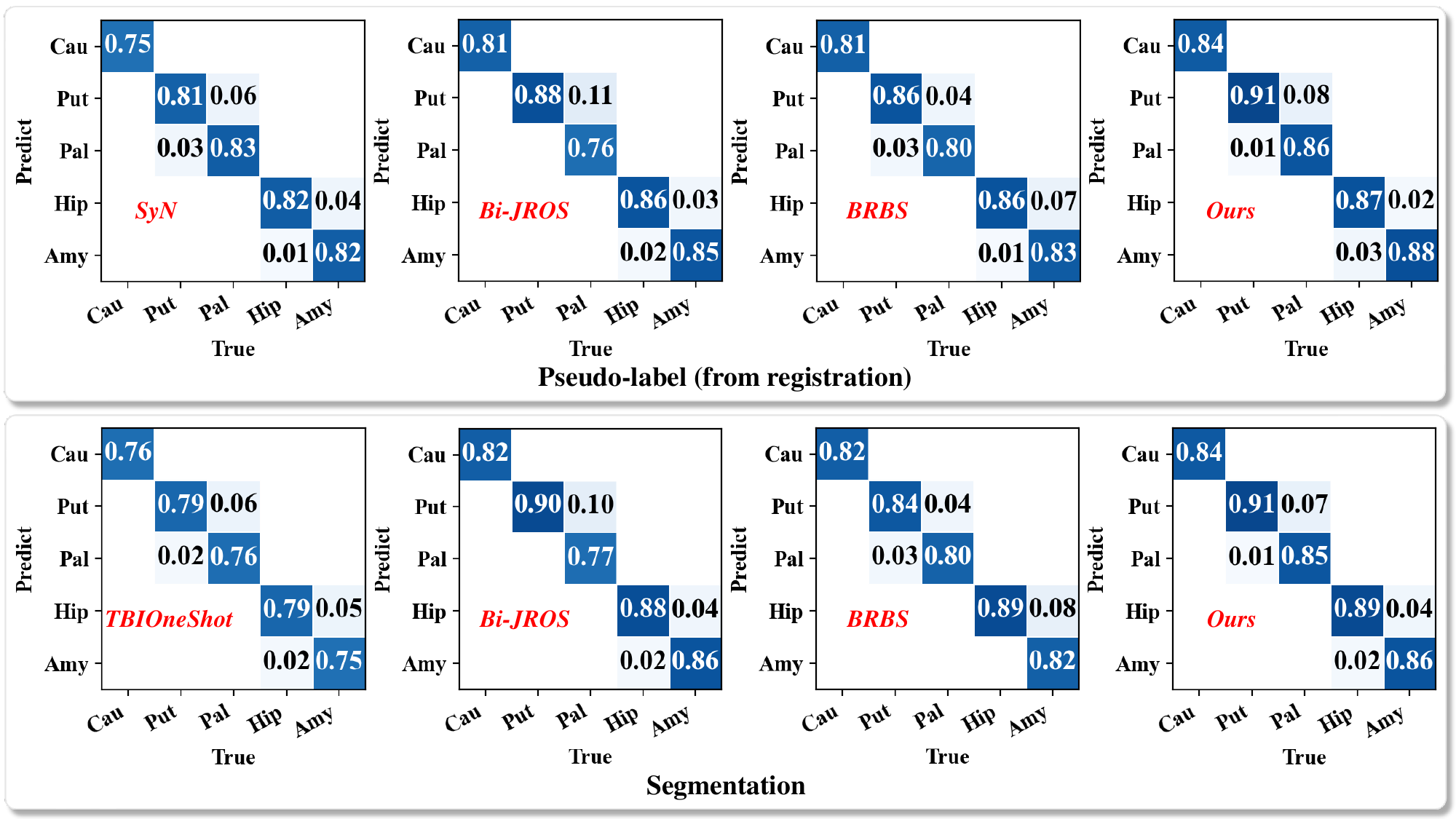}
    \vspace{-0.5cm}
    \caption{Confusion matrices of small anatomical structures (<1\% volume) on the Brain dataset under the 1-shot setting for both registration and segmentation. Pseudo-labels are derived from registration.}
    \vspace{-0.5cm}
    \Description{Confusion matrices comparing registration and segmentation performance on small anatomical structures with volume ratio below one percent.}
    \label{fig:small}
\end{figure}

\noindent\textbf{Qualitative Results.}
As illustrated in Fig.~\ref{fig:seg}, H²AL produces segmentation results that are most consistent with the ground truth across both datasets, particularly for small and structurally complex anatomical regions. Fully supervised methods are limited by scarce annotations, while methods relying solely on Euclidean representations tend to over-smooth fine boundaries due to insufficient modeling of hierarchical context.
In contrast, H²AL effectively captures both local and global anatomical structures by jointly leveraging Euclidean and hyperbolic spaces, resulting in more precise and structurally consistent segmentation.

The qualitative registration results in Fig.~\ref{fig:reg} further demonstrate that H²AL achieves more accurate alignment, especially in regions with ambiguous boundaries and small anatomical structures. This improvement stems from its ability to model hierarchical relationships, leading to better structural correspondence and higher-quality pseudo-labels that further benefit segmentation. As a result, H²AL produces anatomically more consistent registration results.

% These qualitative observations are consistent with the quantitative results, further validating the effectiveness of H²AL in both segmentation and registration tasks.

\begin{table}[t]
\centering
\caption{Registration failure case analysis on Brain (Image A, Image C) and Cardiac (Image B, Image D) datasets. "$\downarrow$" indicates the decrease in Dice compared to the average Dice over the test set, while "$\Uparrow$" indicates the increase.}
\vspace{-0.2cm}
\resizebox{\linewidth}{!}{%
\begin{tabular}{c|ccc|ccc|ccc}
\toprule
\multicolumn{10}{c}{
  \textcolor{lightpink}{\rule{10pt}{7pt}} : Best performance \quad
  \textcolor{lightblue}{\rule{10pt}{7pt}} : Second best performance
} \\
\bottomrule
\toprule
\multirow{2}{*}{\textbf{Cases}}
& \multicolumn{3}{c|}{\textbf{BRBS}} 
& \multicolumn{3}{c|}{\textbf{Bi-JROS}} 
& \multicolumn{3}{c}{\textbf{Ours}} \\
\cmidrule(lr){2-4} \cmidrule(lr){5-7} \cmidrule(lr){8-10}
 & MAE & Dice & \textbf{$\Delta$Dice} 
 & MAE & Dice & \textbf{$\Delta$Dice} 
 & MAE & Dice & \textbf{$\Delta$Dice} \\
\midrule

Image A 
& 0.0582 & 83.45 & \cellcolor{lightblue}\textbf{0.81$\downarrow$} 
& 0.0435 & 82.42 & \textbf{1.13$\downarrow$} 
& 0.0545 & 84.47 & \cellcolor{lightpink}\textbf{0.72$\Uparrow$} \\
\midrule

Image B 
& 0.0326 & 86.40 & \cellcolor{lightblue}\textbf{0.92$\downarrow$} 
& 0.0260 & 85.68 & \textbf{4.13$\downarrow$} 
& 0.0265 & 90.31 & \cellcolor{lightpink}\textbf{0.07$\Uparrow$} \\
\midrule

Image C 
& 0.0481 & 83.06 & \cellcolor{lightblue}\textbf{1.20$\downarrow$} 
& 0.0403 & 82.13 & \textbf{1.42$\downarrow$} 
& 0.0500 & 84.36 & \cellcolor{lightpink}\textbf{0.83$\downarrow$} \\
\midrule

Image D 
& 0.0182 & 84.23 & \cellcolor{lightblue}\textbf{6.15$\downarrow$} 
& 0.0213 & 78.16 & \textbf{11.65$\downarrow$} 
& 0.0260 & 85.38 & \cellcolor{lightpink}\textbf{4.86$\downarrow$} \\
\bottomrule
\end{tabular}
}
\label{table:reg_failure}
\end{table}

\begin{figure}[t]
    \centering
    \includegraphics[width=\linewidth]{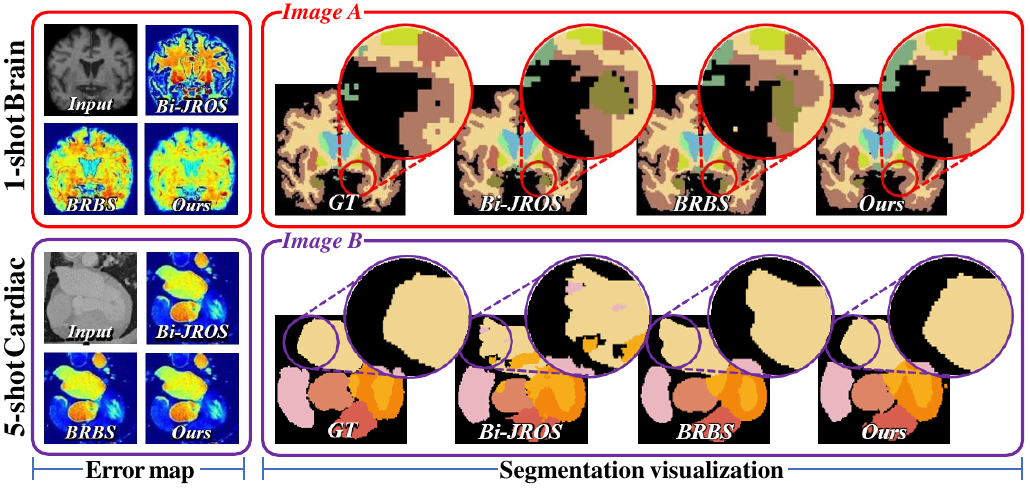}
    \vspace{-0.6cm}
    \caption{
        Visualization of the most severe registration failure cases from each dataset, with Image A from Brain and \mbox{Image B from Cardiac}.
        % “Image A” and “Image B” correspond to the cases shown in Table~\ref{table:reg_failure}.
    }
    \vspace{-0.5cm}
    \Description{
        Registration failure case analysis. $\Delta$Dice denotes the Dice change of the current test image relative to the average of overall test dataset performance.
    }
    \label{fig:reg_failure}
\end{figure}

\subsection{Challenging-case Analysis}
\textbf{Performance on Small Anatomical Structures.}
To further evaluate performance on small anatomical regions, we present confusion matrices of representative structures in Fig.~\ref{fig:small}.
Our method shows higher agreement with the ground truth than competing approaches, indicating more accurate segmentation and improved registration performance on small structures.
We attribute this improvement to two key factors. First, the strong registration capability enables high-quality pseudo-label generation. Second, joint modeling of hierarchical structures facilitates effective integration of structural priors.
As shown in Fig.~\ref{fig:seg} and Fig.~\ref{fig:reg}, our method better preserves fine structural boundaries in small regions while avoiding the over-smoothing and misalignment observed in competing methods, demonstrating its effectiveness in capturing fine-grained details with strong structural consistency.

\noindent\textbf{Robustness under Registration Failure.}
A practical concern for registration-based FMIS is whether segmentation degrades under registration failure. To investigate this, we select the two most severe failure cases per dataset based on MAE error maps, corresponding to the largest discrepancies between warped source and target images. Table~\ref{table:reg_failure} reports Dice variations under these extreme scenarios, compared with state-of-the-art methods, along with qualitative results of the worst case per dataset in Fig.~\ref{fig:reg_failure}. Despite substantial registration errors, H$^2$AL consistently achieves superior segmentation performance and even yields \textbf{positive Dice gains in two of four cases}. This robustness stems from the hierarchical modeling strategy, which first aligns parent structures to establish stable global anchors and then refines child structures, mitigating error propagation caused by misalignment.

\begin{table}[t]
\centering
\caption{Ablation study results on the Brain dataset under the Atlas setting. “E” and “H” denote Euclidean and hyperbolic spaces, respectively.}
\vspace{-0.2cm}
\label{tab:ablation_table}
\resizebox{\linewidth}{!}{%
\begin{tabular}{c|cc|c|ccc|ccc}
\toprule[0.3mm]
\multicolumn{10}{c}{
  \coloredcircle{mypurple}{S} : Small structure \quad
  \coloredcircle{myblue}{L} : Large structure \quad
  \coloredcircle{mygreen}{T} : Total
} \\
\bottomrule
\toprule
\multirow{2}{*}{\textbf{Methods}} &
\multicolumn{2}{c|}{\textbf{Setting}} &
\multirow{2}{*}{\textbf{Space}} &
\multicolumn{3}{c|}{\textbf{Reg-Dice (\%)}} &
\multicolumn{3}{c}{\textbf{Seg-Dice (\%)}} \\
\cmidrule(lr){2-3}\cmidrule(lr){5-7}\cmidrule(lr){8-10}
& \textbf{Reg} & \textbf{Seg} & 
& \textbf{\coloredcircle{mypurple}{S}} & \textbf{\coloredcircle{myblue}{L}} & \textbf{\coloredcircle{mygreen}{T}} 
& \textbf{\coloredcircle{mypurple}{S}} & \textbf{\coloredcircle{myblue}{L}} & \textbf{\coloredcircle{mygreen}{T}} \\
\midrule

\textbf{baseline}
  & - & - & E
  & 77.18 & 78.53 & 78.01
  & 78.06 & 81.29 & 80.05 \\
\midrule

\textbf{+GA}
  & - & - & E
  & \textbf{77.76}
  & \textbf{79.10}
  & \textbf{78.58}
  & \textbf{78.63}
  & \textbf{82.05}
  & \textbf{80.73} \\
\midrule

\multirow{3}{*}{\textbf{+Hyperbolic}}
  & \cellcolor{lightblue}\cmark & \cellcolor{lightblue}\xmark & H
  & 78.06
  & 79.46
  & 78.92
  & 79.15
  & 81.90
  & 80.84 \\
  
  & \cellcolor{deepblue}\xmark & \cellcolor{deepblue}\cmark & H
  & 77.88
  & 79.16
  & 78.66
  & 79.29
  & 81.92
  & 80.91 \\
  
  & \cellcolor{lightpink}\cmark & \cellcolor{lightpink}\cmark & H
  & \textbf{78.28}
  & \textbf{79.52}
  & \textbf{79.04}
  & \textbf{79.34}
  & \textbf{81.96}
  & \textbf{80.95} \\
\midrule

\multirow{3}{*}{\textbf{+TSHCL}}
  & \cellcolor{lightblue}\cmark & \cellcolor{lightblue}\xmark & E\&H
  & 79.52
  & 80.67
  & 80.23
  & 79.37
  & 82.15
  & 81.08 \\
  
  & \cellcolor{deepblue}\xmark & \cellcolor{deepblue}\cmark & E\&H
  & 78.54
  & 79.71
  & 79.26
  & 80.64
  & 83.41
  & 82.34 \\
  
  & \cellcolor{lightpink}\cmark & \cellcolor{lightpink}\cmark & E\&H
  &\textbf{79.59}
  & \textbf{81.89}
  & \textbf{80.75}
  & \textbf{81.07}
  & \textbf{83.69}
  & \textbf{82.69} \\
\midrule

\multirow{3}{*}{\textbf{+GIB}}
  & \cellcolor{lightblue}\cmark & \cellcolor{lightblue}\xmark & E\&H
  & 81.16
  & 82.04
  & 81.70
  & 81.30
  & 84.16
  & 83.05 \\
  
  & \cellcolor{deepblue}\xmark & \cellcolor{deepblue}\cmark & E\&H
  & 80.45
  & 81.31
  & 80.98
  & 82.29
  & 85.01
  & 83.96 \\
  
  & \cellcolor{lightpink}\cmark & \cellcolor{lightpink}\cmark & E\&H
  & \textbf{81.21}
  & \textbf{82.42}
  & \textbf{81.96}
  & \textbf{82.37}
  & \textbf{85.30}
  & \textbf{84.17} \\
\bottomrule
\end{tabular}
}
\end{table}

\begin{figure}[t]
    \centering
    \includegraphics[width=\linewidth]{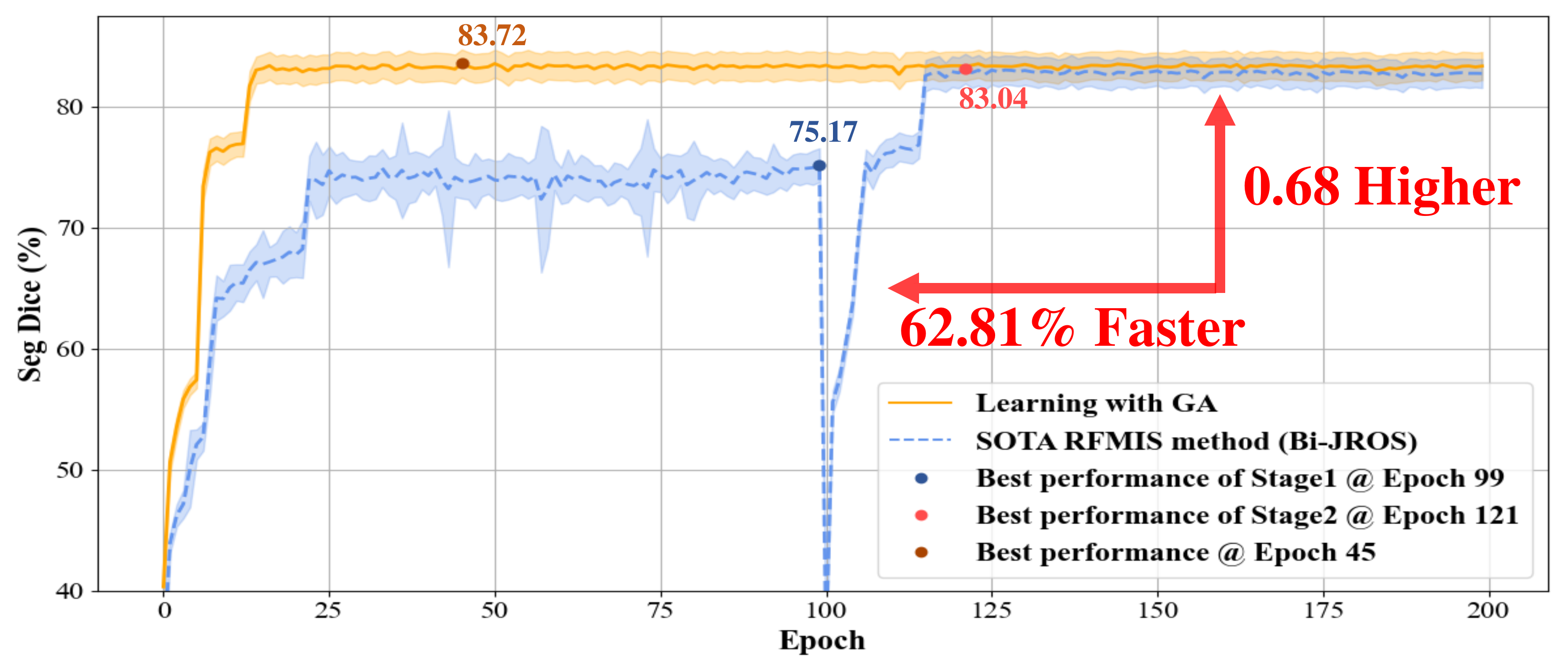}
    \vspace{-0.7cm}
    \caption{Growth curves of segmentation Dice for the SOTA RFMIS~\cite{fan2024bi} training strategy and our proposed strategy on the Brain dataset (Atlas setting).}
    \Description{Growth curves comparing segmentation Dice during training for the SOTA RFMIS strategy and the proposed strategy on the Brain dataset under the atlas setting.}
    \vspace{-0.3cm}

    \label{fig:ours-faster}
\end{figure}

\subsection{Mechanism Analysis of H$^2$AL}

\textbf{Why Infusing Hyperbolic into Euclidean Matters?}

Euclidean and hyperbolic geometries provide complementary inductive biases for representation learning. As shown in Table~\ref{tab:ablation_table}, Euclidean modeling already provides a strong baseline, while hyperbolic modeling further improves performance, particularly in registration, due to its ability to capture hierarchical structures.
However, each geometry has inherent limitations. As illustrated in Fig.~\ref{fig:enter-label}, Euclidean embeddings exhibit compact clusters but weak global structure, whereas hyperbolic embeddings capture clear hierarchies at the cost of compactness. In contrast, our method integrates both spaces via H2I, achieving both compact intra-class clustering and well-structured inter-class separation.
This balance is especially critical for small anatomical structures. By enabling global hierarchical anatomical consistency and enhanced anatomical discrimination, our method better captures them, leading to improved segmentation performance.

\noindent\textbf{Why Gradient Aggregation Matters?}

Gradient aggregation (GA) plays a key role in stabilizing optimization across heterogeneous geometric spaces. As shown in Fig.~\ref{fig:ours-faster}, it accelerates convergence and improves the final performance, as separately induced task updates may lead to inconsistent optimization dynamics and unstable training. To overcome this issue, we propose a GA strategy that integrates task-specific gradients enriched with semantic and hierarchical cues to jointly optimize the shared encoder for registration and segmentation. Such a design encourages more coordinated dual-task learning, alleviates optimization inconsistency, and enhances training stability. This is well aligned with the motivation in Fig.~\ref{fig:fig3}, underscoring the necessity of coordinated optimization in multi-space learning.

% \begin{figure}[t]
%     \makebox[\textwidth][l]{
%         \includegraphics[width=\linewidth]{res/hyper_ablation.pdf}
%     }
%     \vspace{-0.6cm}
%     \caption{Hyperparameter sensitivity and convergence stability analyses.}
%     \vspace{-0.7cm}
%     \label{fig:hyper_ablation}
% \end{figure}

\subsection{Ablation Study}

Table~\ref{tab:ablation_table} presents the component-wise ablation results. The baseline is constructed by pre-training a shared encoder, then freezing it and alternately optimizing the registration and segmentation decoders.

\noindent\textbf{Effect of GA.}
After introducing the GA strategy, the model performs end-to-end joint optimization within a single stage. Compared with the baseline, both registration and segmentation show consistent performance improvements. These results indicate that jointly optimizing the two tasks leads to more effective feature learning, suggesting that the two objectives provide complementary supervisory signals rather than competing ones.

\noindent\textbf{Effect of H2I.}
Table~\ref{tab:ablation_table} further reports the performance across anatomical structures of different scales. Incorporating \textbf{TSHCL} and \textbf{GIB} into H2I brings consistent gains in both registration and segmentation, especially for small-scale structures. This verifies that H2I can effectively bridge the complementary strengths of Euclidean and hyperbolic spaces by preserving the semantic richness of Euclidean representations while injecting the hierarchy-aware priors encoded in hyperbolic geometry. Such a design enables the model to learn more discriminative and anatomically structured features, thereby simultaneously improving deformation quality and segmentation performance.

\section{Conclusion}

In this paper, we propose H²AL that enhances both deformation plausibility and anatomical discrimination for dual-task learning. Specifically, we introduce a Hyperbolic Hierarchy-aware Infusion (H2I) module, which leverages the hierarchical capacity of hyperbolic space to learn precise hierarchy-aware representations and injects such priors into Euclidean space via a gated infusion block. Then, we develop an end-to-end training strategy with Gradient Aggregation (GA) to ensure consistent optimization between registration and segmentation. Extensive experiments on brain and cardiac datasets demonstrate the superior performance of our method, especially for small anatomical structures. \emph{More broadly}, we believe that jointly modeling Euclidean semantics, hyperbolic hierarchy, and cross-task gradient interaction provides a promising and extensible direction for future medical image analysis.

\begin{acks}
This work was supported by the
\grantsponsor{GS1}{National Natural Science Foundation of China}
{https://doi.org/10.13039/501100001809}
under Grant No.~\grantnum{GS1}{61733002}.
\end{acks}

\bibliographystyle{ACM-Reference-Format}
\bibliography{ref}

\end{document}